\documentclass[conference]{IEEEtran}

\usepackage{times}
\usepackage{latexsym}
\usepackage{amssymb}
\usepackage[T1]{fontenc}

\usepackage[utf8]{inputenc}

\usepackage{microtype}

\usepackage{ragged2e}
\usepackage{paralist}
\usepackage{booktabs}
\usepackage{enumitem}
\usepackage{graphicx}
\usepackage{array}
\usepackage{url}
\usepackage{tabularx}
\usepackage[export]{adjustbox}
\usepackage{multirow}
\usepackage{tikz}
\usepackage{tcolorbox}

\newcolumntype{P}[1]{>{\centering\arraybackslash}p{#1}}
\newcolumntype{L}[1]{>{\raggedright\arraybackslash}p{#1}}
\newcolumntype{C}[1]{>{\centering\arraybackslash}p{#1}}
\newcolumntype{R}[1]{>{\raggedleft\arraybackslash}p{#1}}

\newcolumntype{Y}{>{\RaggedRight\arraybackslash}X}

\newcommand{\tweetcase}[4][0.78in]{%
\begin{minipage}[t]{\linewidth}
\vspace{0pt}
\begin{minipage}[t]{0.62\linewidth}
\vspace{0pt}
#2
\end{minipage}\hfill
\begin{minipage}[t]{0.34\linewidth}
\vspace{0pt}
\centering
\includegraphics[width=\linewidth,height=#1,keepaspectratio]{#3}
\end{minipage}\par
\vspace{4pt}
#4
\end{minipage}%
}

\begin{document}
\title{Leaving, Staying, or Living There? \\
Duration Commonsense Reasoning with Human-LLM Co-annotations
}


\author{\IEEEauthorblockN{Chenyang Yu\textsuperscript{1}, Zhaomin Xiao\textsuperscript{1}, Yan Huang\textsuperscript{1}, Eduardo Blanco\textsuperscript{2}}

\IEEEauthorblockA{\textsuperscript{1}University of North Texas \textsuperscript{2}University of Arizona}
\IEEEauthorblockA{chenyangyu@my.unt.edu, zhaominxiao@my.unt.edu, yan.huang@unt.edu, eduardoblanco@arizona.edu. \\ }
}

\maketitle

\begin{abstract}

Gardening in the backyard implies living there, while checking into an Airbnb more likely means staying for a few days. Such duration judgments are important for understanding human mobility, travel behavior, and social context. However, duration is rarely stated explicitly in social media posts and must instead be inferred from noisy textual, visual, and contextual cues. In this paper, we investigate whether vision-language models possess such duration commonsense. We create a new corpus of geotagged tweets co-annotated by humans and LLMs to serve as a testbed for this task. We find that state-of-the-art models struggle with duration inference, even when guided by multi-turn Chain-of-Thought prompting. Error analyses further reveal that models often fail to integrate critical contextual information, or they overly rely on prominent visual details while overlooking subtle hashtags and semantic nuances. These findings underscore the significant challenges remaining in context-aware multimodal understanding.

\end{abstract}
\begin{IEEEkeywords}
Duration commonsense reasoning, vision-language models, social media, human-LLM co-annotation, multimodal understanding.
\end{IEEEkeywords}
\section{Introduction}
\label{section:introduction}

Duration commonsense, the ability to infer how long human activities last, is fundamental to human intelligence and allows us to navigate temporal contexts. Extracting duration commonsense from social media benefits applications such as transportation planning and personalized recommendations. For example, inferring how long users stay at checked-in locations helps estimate travel demand. 
However, because this inference requires interpreting nuanced social, visual, and textual cues, developing computational models to replicate this human ability remains challenging.

\begin{figure}
    \includegraphics[width=\linewidth]{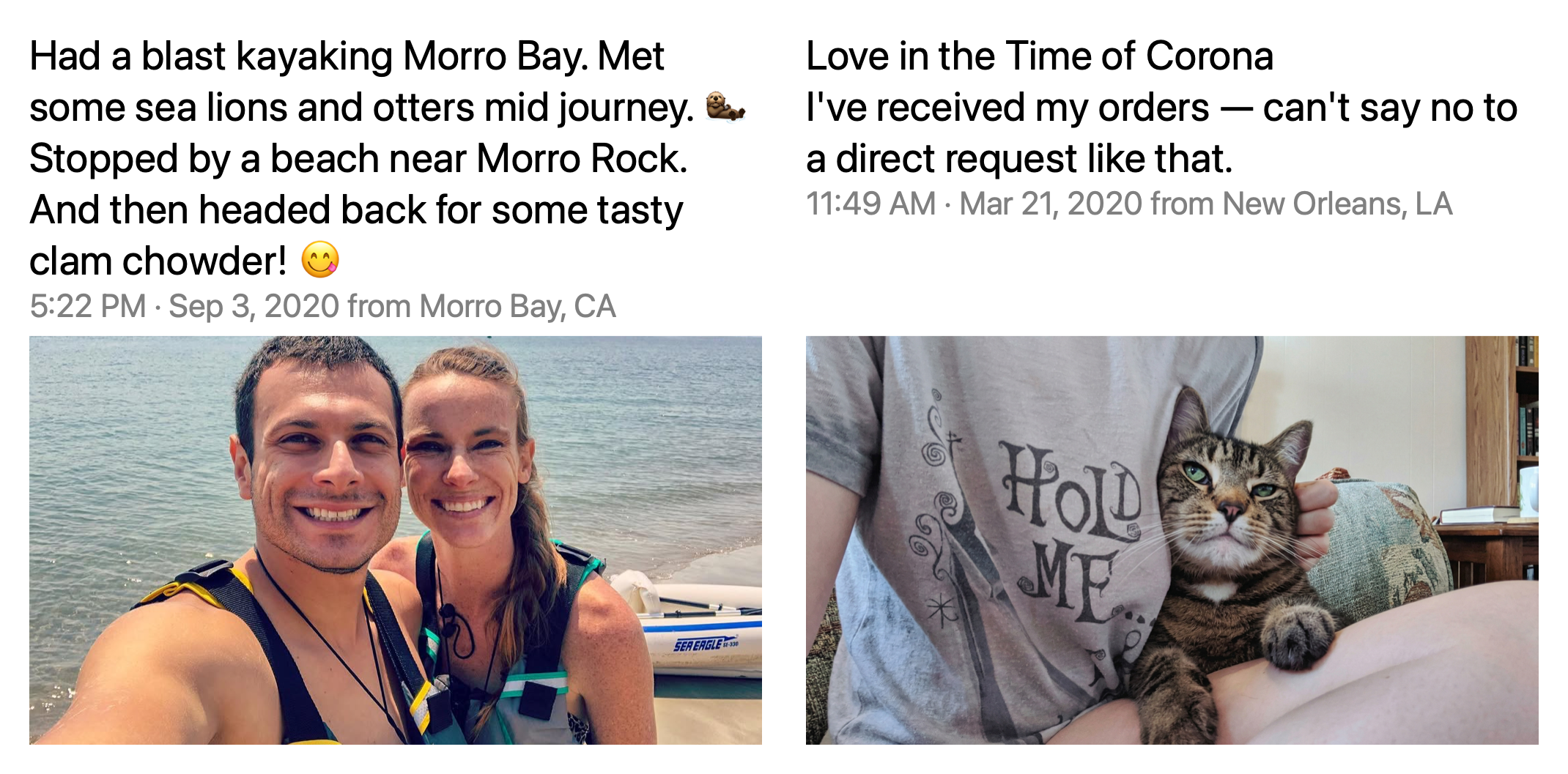}
    \caption{
            Examples of tweets in which the author will stay at the geotagged location for a short (left) or long (right) time.
            The author of the left tweet will stay at \textit{Morro Bay} only briefly: the text recounts kayaking before heading back, and the image shows the author in a kayak during a day trip.
            The author of the right tweet will stay at \textit{New Orleans} for a long time; the ``Love in the Time of Corona'' post, indoor setting, and pet suggest the author is sheltering at home during the COVID-19 lockdown.
        }
        \label{fig:motivation}
\end{figure}

Consider the tweets in Figure~\ref{fig:motivation}. 
The author of the left tweet will likely stay briefly in \textit{Morro Bay}: the text recounts kayaking before heading back, and the image shows the author in a kayak with life jackets during a day trip.
Conversely, the right tweet suggests a stay of months or years in \textit{New Orleans}; the indoor setting, pet, and references to sheltering in place indicate the author is living there during the lockdown.


Prior duration research focuses on event duration inference \cite{gusev-etal-2011-using,leeuwenberg-moens-2018-temporal}, predicting how long events last rather than how long individuals participate. While some work extracts human-centric duration from formal texts \cite{vempala-blanco:2016:ACL,chinnappa-etal-2020-beyond}, social media shares human experiences through noisy, multimodal posts where duration is rarely explicit. Inferring stay duration at a specific location requires reasoning across informal text, hashtags, and visual elements, a socially nuanced task that remains largely underexplored.

In this paper, we focus on this gap by investigating the problem of using noisy user-generated content to determine how long people will stay at the places they tag in their tweets posted on X (formerly known as Twitter). 
People talk about their vacation trips, work, and opinions in tweets. Tweets contain not only text (e.g., hashtags, emojis, and URLs), 
but also multimedia content (e.g., images and videos) and geotags. Indeed, 42\% of tweets have images~\cite{TwitterImgs},
Tweets with images receive 150\% more retweets than tweets without them~\cite{imageretweet}.
On the contrary, geotags are not consistently available because sharing location information when tweeting is not enabled by default due to privacy concerns.

The main contributions of this paper are:
\begin{inparaenum}[(a\upshape)]
\item~ We {\bf formulate a novel and socially grounded problem}: inferring duration commonsense about people from text and images in social media. The task is inherently challenging, requiring nuanced interpretation of social context that even humans find difficult;
\item~ We construct a {\bf Human–LLM co-annotated Corpus} to address the challenges of annotating socially complex phenomena. By combining human judgment with the consistency and scalability of large language models, we create {\bf a corpus of 4,312 geotagged tweets} with duration annotations indicating how long tweet authors are likely to stay at the locations mentioned in their posts\footnote{Corpus and code available at \url{https://anonymous.4open.science/r/loc_tmp3-2327/}};
\item~We systematically evaluate the {\bf pretrained language models (PTLMs), text-only LLMs, and VLMs}. Our results show that while multi-turn Chain-of-Thought prompting and advanced reasoning models significantly improve performance, they struggle with highly ambiguous and culturally nuanced cues;
\item~We perform detailed error analyses, providing {\bf insights} into the fundamental bottlenecks of multimodal duration inference. We find that complex semantic reasoning remains the primary challenge, as models frequently overlook contextual constraints or conflate disjoint visual and textual clues.
\end{inparaenum}

\section{Related Work}
\label{section:related_work}

Duration has been investigated since it was proposed long ago~\cite{10.1162/COLI_a_00075}.
In their pioneer work, they create a corpus consisting of news articles in which events are annotated for estimated bounds on their duration, as well as providing simple machine learning baseline models that take into account word n-grams, syntactic relations, and WordNet hypernyms~\cite{10.1093/ijl/3.4.235}.
Since neural networks started to be adopted in NLP~\cite{socher-etal-2011-semi,chen-manning-2014-fast,kim-2014-convolutional}, many neural network-based models are proposed to predict event durations using formal text, including clinical documents and news articles~\cite{yang-etal-2020-improving,9207839,ma-etal-2021-eventplus,dasgupta2025predicting}.
In the realm of social media data, \cite{williams-katz-2012-extracting} create a dataset of 14 million tweets and build a bootstrapping classifier for the event duration. 
However, the duration of the event differs from the duration of people’s stays.
People may or may not participate in the events they mention in their tweets~\cite{sanagavarapu-vempala-blanco:2017:Short}. 
Even if people participate in the mentioned events, the duration of those events may not correspond to the duration of people’s stays. 
Different from previous efforts aiming at understanding commonsense knowledge about event-centric duration~\cite{zhou-etal-2019-going,zhou-etal-2020-temporal,pereira-etal-2020-adversarial,pereira-etal-2021-alice}, our work presented in this paper focuses on understanding people-centric duration commonsense.
The main difference is that the duration commonsense we focus on in this paper is not centered around events but around people. For example, a tweet like ``Looks like the music festival downtown is absolutely packed this weekend! Everyone's sharing such amazing photos \#FestivalVibes'' discusses an event (the music festival), but it does not imply that the user is participating in it, and we cannot infer how long the user will stay there. In contrast, a tweet like ``Finally made it to the tech conference! Excited to spend the next three days soaking up all the insights and networking \#TechConf2025'' provides clear duration information, as it mentions the user’s plans to stay for three days.

\begin{figure*}[t]
    \small
    \centering
    \begin{tabular}{@{}c@{\hskip 0.2in}@{\hskip 0.2in}c@{\hskip 0.2in}@{\hskip 0.2in}c@{}}
    
        \includegraphics[width=1.8in]{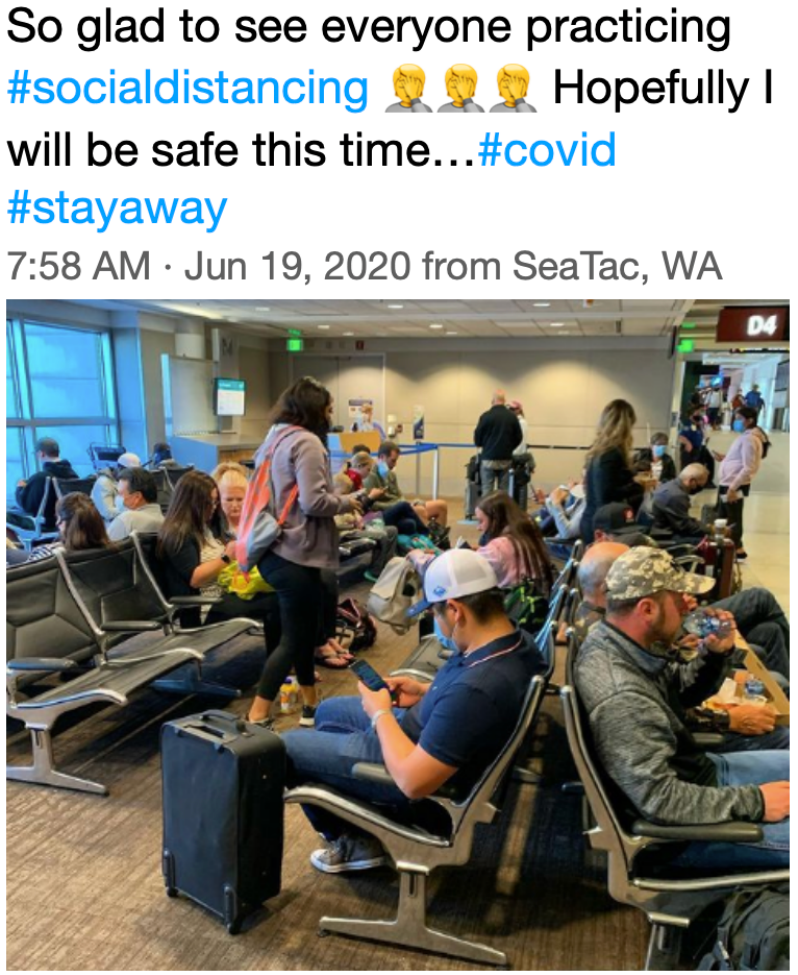}
        & \includegraphics[width=1.8in]{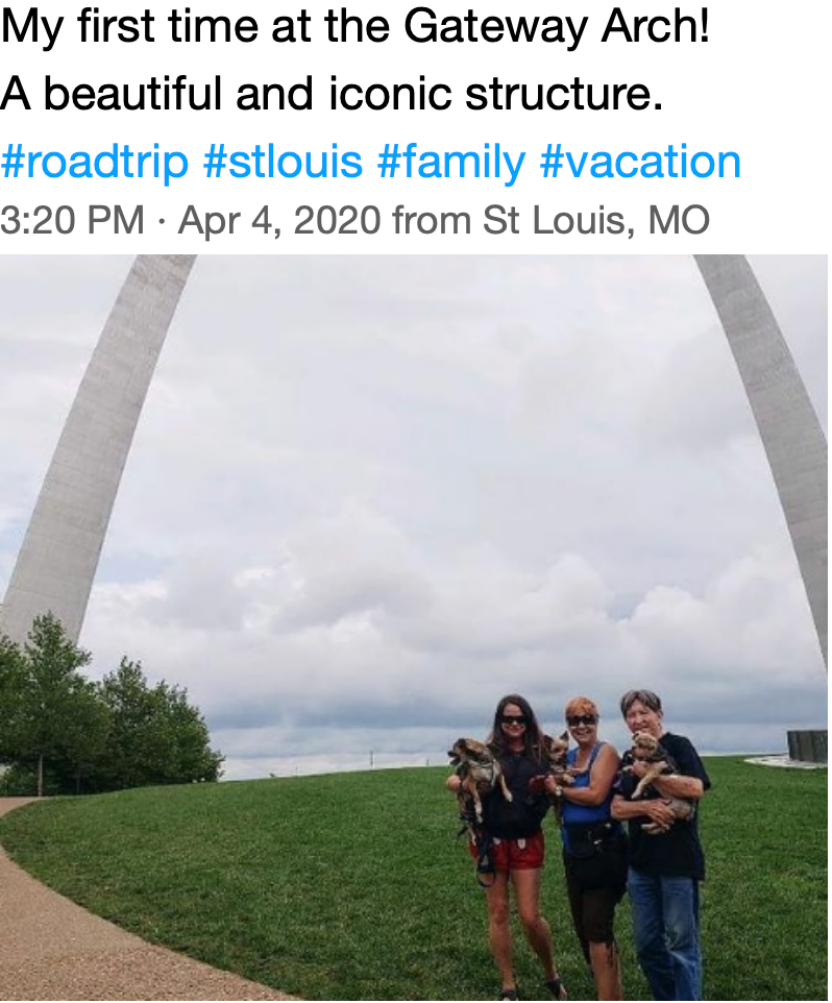}
        & \includegraphics[width=1.8in]{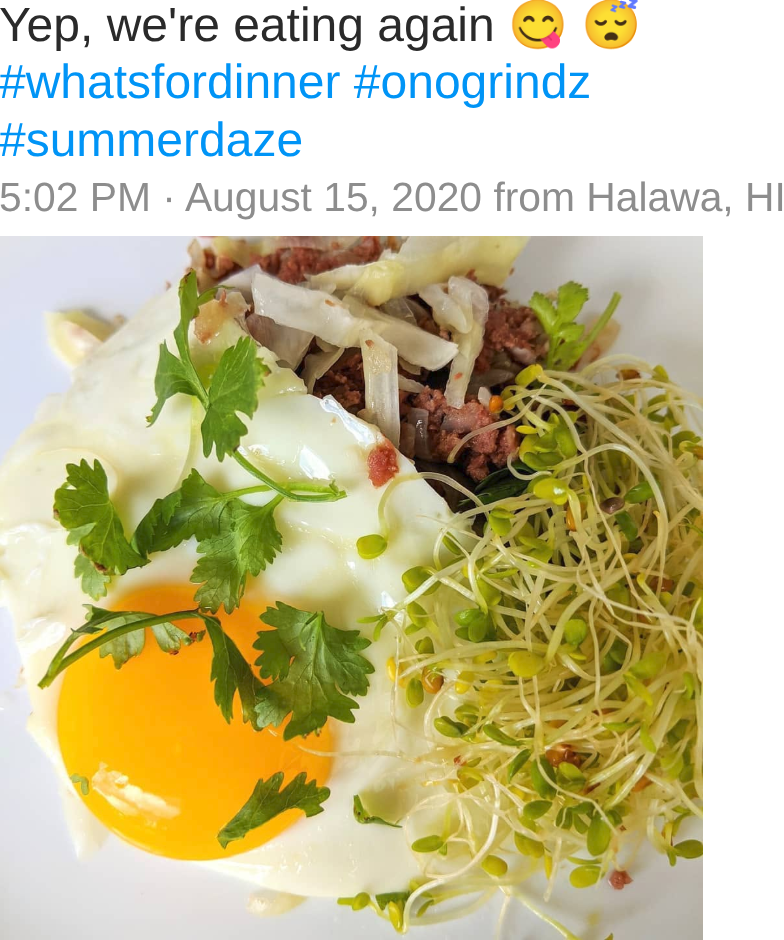} \\ 
        
        \texttt{0-24 hours} 
        & \texttt{1-30 days} 
        & \texttt{more than 30 days} \\ 
        
    \end{tabular}

    \caption{Annotation examples.
        The author of the leftmost tweet is departing \emph{SeaTac} (``Hopefully I will be safe this time'' in the text and the crowded airport waiting area in the image), indicating a stay of \texttt{0-24 hours}.
        The middle tweet exemplifies \texttt{1-30 days}: the author visits the Gateway Arch on a \#roadtrip to \textit{St Louis}, and travelers typically spend less than one week at each stop.
        The author of the rightmost tweet lives in \textit{Halawa}; the ``Yep, we're eating again'' text and a home-cooked meal in the image, together with hashtags such as \#whatsfordinner and \#onogrindz, suggest everyday residence rather than a brief visit, leading to \texttt{more than 30 days}.
    } 
    \label{figure:corpus_example}

\end{figure*}  
There have been few previous works studying duration related to people. 
For instance, \cite{chinnappa-etal-2020-beyond} determine the duration of the possession relations, in which possessions are defined as someone possessing something. 
More related to our work, \cite{vempala_blanco_2021} extract where people are located and for how long, creating a corpus based on OntoNotes~\cite{10.5555/1614049.1614064}, which consists of formal text. 
In contrast, our work is complementary in two ways. 
First, we work with informal text, which makes our task more challenging. 
Second, we consider not only text but also images from which duration commonsense could be inferred.

Annotation of such socially complex task is itself challenging. While crowdsourcing is relatively inexpensive and flexible, the resulting quality can be inconsistent. Recent advances in large language models (LLMs) have made them attractive for annotation. There is a significant body of research explores using LLMs to support or replace human annotators \cite{wang-etal-2021-want-reduce,CrowdLLMLabel}, with promising results across a range of domains. Notably, LLMs have shown strong performance in annotating socially nuanced tasks~\cite{doi:10.1073/pnas.2305016120}, sometimes even surpassing human annotators~\cite{10.1145/3613904.3642834}. These capabilities enable the design of automated annotation pipelines that are accessible without extensive technical expertise, broadening their applicability.
Our work is motivated by these previous efforts. However, to the best of our knowledge, we are the first to study the duration commonsense about people using text and images on social media. 

\section{A Corpus of Duration Commonsense in Tweets }
\label{section:corpus_building}

We create a new corpus as there is not an existing dataset. To capture duration commonsense as understood by both humans and large language models (LLMs), we adopt a hybrid annotation strategy that integrates crowdworker and LLM annotations. Human annotators contribute socially grounded judgments shaped by real-world experience, while LLMs provide scalable and consistent interpretations, having shown strong performance on socially nuanced tasks. Combining both sources allows us to leverage their complementary strengths: human contextual sensitivity and LLM consistency. We then introduce a process for combining these human and LLM labels into a cohesive and high-quality dataset.

\subsection{Collecting Tweets}
\label{section:collecting_tweets}

We use \textsc{Geo}\textsc{Cov19}\textsc{Tweets}~\cite{fpsb-jz61-20}, containing approximately 420,000 tweets. We selected this dataset because its metadata provides precise geotags and it captures a wide range of stay durations—from brief essential outings to prolonged home stays. Although pandemic-era data exhibits behavioral biases, the underlying multimodal reasoning required to infer duration remains generalizable. Figure~\ref{figure:corpus_example} illustrates annotation examples spanning all three duration labels, from a traveler leaving \emph{SeaTac} to a \#roadtrip stop in \textit{St Louis} and everyday home life in \textit{Halawa}. We downloaded the full metadata for the provided tweet IDs via the X Developer API.

\paragraph{Tweets Filtering}
We filtered the dataset using three criteria:
First, we retained only tweets with US city-level geotags, excluding broader state-level or narrower point-of-interest tags.
Second, each tweet must contain both text and images.
Finally, we excluded corporate or organizational accounts using keyword matching on usernames (e.g., \textit{daily}, \textit{sales}) and poster-like image detection via CRAFT~\cite{DBLP:conf/cvpr/BaekLHYL19}. 

Additionally, we removed tweets containing explicit duration phrases (e.g., \textit{5 days}, \textit{one year}) to ensure the task evaluates duration inference rather than explicit extraction.


\subsection{Annotation Guidelines}
\label{section:annotation_guidelines}

We ask both crowdworders and LLM to annotate:
\begin{itemize}[nosep]
    \item {Amazon Mechanical Turk (AMT): } We utilized the Amazon Mechanical Turk platform to obtain annotations from crowd-workers.
    \item {LLM Annotation: } Six Large Language Models (LLMs), Qwen 2.5~\cite{bai2025qwen25vltechnicalreport}, GPT5-mini \cite{GPT5mini}, Grok4 \cite{grok}, Claude 3 haiku, Gemini 2.5 flash and Gemini 2.5 pro \cite{gemini} were employed to generate annotations.
\end{itemize}
For crowdworkers, we present each tweet with text and image(s) and ask them a straightforward question: ``How long will the author of the tweet stay in the \emph{city} continuously after publishing this tweet?'' The \emph{city} in question is the geotagged city of the tweet.
For instance, given a tweet with the text ``\textit{I am gonna join abortion law protest in Dallas tomorrow}'', its associated image, and a geotag of \emph{Plano, TX}, we can infer that the author of the tweet will stay in Plano for less than 24 hours and then go to Dallas.
Crowdworkers choose among four options corresponding to \emph{leaving}, \emph{staying}, \emph{living}, and \emph{can not determine}. 
Approximately 2,500 annotators participated in our crowdsourcing efforts. Collecting annotations from a diverse pool of crowdworkers was essential because duration interpretation varies across age groups and educational backgrounds. To ensure high-quality labels for this ambiguous task, we implemented rigorous quality control. We discarded approximately 30\% of the annotations collected through Amazon Mechanical Turk (AMT) because they failed sanity checks or exhibited random clicking behavior. We compensated crowdworkers above the United States federal minimum wage of \$7.25 per hour, adhering to platform guidelines. After adjudication and filtering, each tweet in our corpus retains an average of 2.5 high-quality annotations.

For LLMs, we use a task-specific system prompt: ``You are a helpful assistant to infer the duration of human activities.'' Each tweet, along with text and image(s), is presented, and the same question with four options is posed to the model. 
Detailed examples of the prompt can be found in Table~\ref{fig:prompt_examples}.




\subsection{Annotation Quality Control of Crowdworkers and LLMs}
\label{section:annotation_quality}
Ensuring annotation quality requires rigorous control for both human and LLM annotators. For crowdworkers, we implement two initial filters based on pilot observations: we recruit only workers with historical approval rates above 95\%, and we enforce a minimum completion time of 5 seconds per Human Intelligence Task (HIT).

We collect five human annotations per HIT and filter unreliable responses to achieve \emph{substantial} inter-annotator agreement. We quantify reliability using Krippendorff’s $\alpha$~\cite{krippendorff1computing}, where $\alpha = 0$ indicates chance-level agreement and $\alpha = 1$ indicates perfect agreement. Values of $\alpha \ge 0.6$ are considered substantial. We repeat the following steps for each batch until $\alpha \geq 0.6$:
\begin{compactenum}
    \item Calculate the competence score for each crowdworker using Multi-Annotator Competence Estimation (MACE)~\cite{hovy-etal-2013-learning}.
    \item Discard all annotations from the crowdworker with the lowest competence score.
    \item Recompute Krippendorff’s $\alpha$ on the remaining annotations.
\end{compactenum}
This iterative process yields a final average Krippendorff’s $\alpha$ of 0.68 across all human annotations. After filtering, we derive the aggregated human labels using MACE to weight the remaining annotators by their inferred competence, avoiding the limitations of simple majority voting.

For LLM annotators, we assess inter-rater agreement using Fleiss’ Kappa (FK)~\cite{Fleiss1971MeasuringNS}. Table~\ref{tab:kappa} reports the $\kappa$ values for each model. We treat models with $\kappa < 0.2$ (Qwen 2.5 and GPT5-mini) as low-quality annotators and exclude them. We then derive the final adjudicated LLM labels by applying MACE to the predictions of the remaining four models.

\begin{table}[h]
\centering
\caption{Inter-model agreement measured using Cohen’s Kappa.}
\label{tab:kappa}
\resizebox{\linewidth}{!}{
\begin{tabular}{lcc}
\toprule
\textbf{Model} & \textbf{Kappa Value} & \textbf{Interpretation} \\
\midrule
Qwen 2.5 & 0.1659 & Moderate agreement \\
GPT5-mini & 0.1892 & Moderate agreement \\
Grok 4 & 0.2985 & Fair agreement \\
Claude 3 haiku & 0.3256 & Fair agreement \\
Gemini 2.5 flash & 0.5302 & Substantial agreement \\ 
Gemini 2.5 pro & 0.5572 & Substantial agreement \\
\bottomrule
\end{tabular}
}
\end{table}

\subsection{LLM and Human Annotation Alignment}
To construct a reliable corpus, we must resolve conflicts between human and LLM consistency. Table~\ref{tab:pair_distribution} presents the joint distribution of human and LLM predictions. The highest agreement occurs for the \texttt{more than 30 days} category, demonstrating that both annotator types share baseline knowledge regarding prolonged quarantine stays.


\begin{table}[h]
\centering
\caption{Confusion matrix of duration labels assigned by human and LLM annotators. Values are percentages. Diagonal cells (in \textbf{bold}) indicate consensus labels between annotators.}
\label{tab:pair_distribution}
\resizebox{\linewidth}{!}{  
\begin{tabular}{lccc}
\toprule
\multirow{2}{*}{\textbf{LLM}} 
& \multicolumn{3}{c}{\textbf{Human}} \\
\cmidrule(lr){2-4}
& 0-24 hours 
& 1-30 days 
& more than 30 days \\
\midrule
0-24 hours  & \textbf{2.00}  & 0.73 & 18.36 \\
1-30 days   & 4.57 & \textbf{2.65}  & 5.40 \\
more than 30 days   & 32.23 & 2.37 & \textbf{31.69} \\
\bottomrule
\end{tabular}
}
\end{table}

\begin{table*}[t]
\centering
\small
\setlength{\tabcolsep}{5pt}
\renewcommand{\arraystretch}{1.08}
\caption{Common system errors made by LLM and human annotators.}
\label{tab:system_error_merged}
\begin{tabularx}{\textwidth}{@{}YYYY@{}}
\toprule
\textbf{R1: COVID-19 quarantine duration commonsense} &
\textbf{R2: Duration not calculated at the city level} &
\textbf{R3: Tweet intent unclear} &
\textbf{R4: Author residency status unclear} \\
\midrule

\tweetcase{
Quarantine Birthday for me. \#lockdown \#coronavirus \#carrotcake \#nyc \#manhattan \#cumplea
}{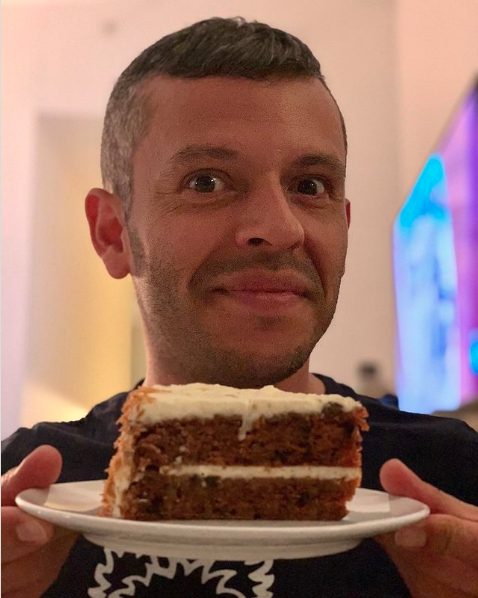}{
3:55 AM Apr 28, 2020 from Manhattan, NY
}
&
\tweetcase{
VVBS buying @tahlequahumc \#vvbs \#vbs2020 @ Tahlequah United Methodist Church
}{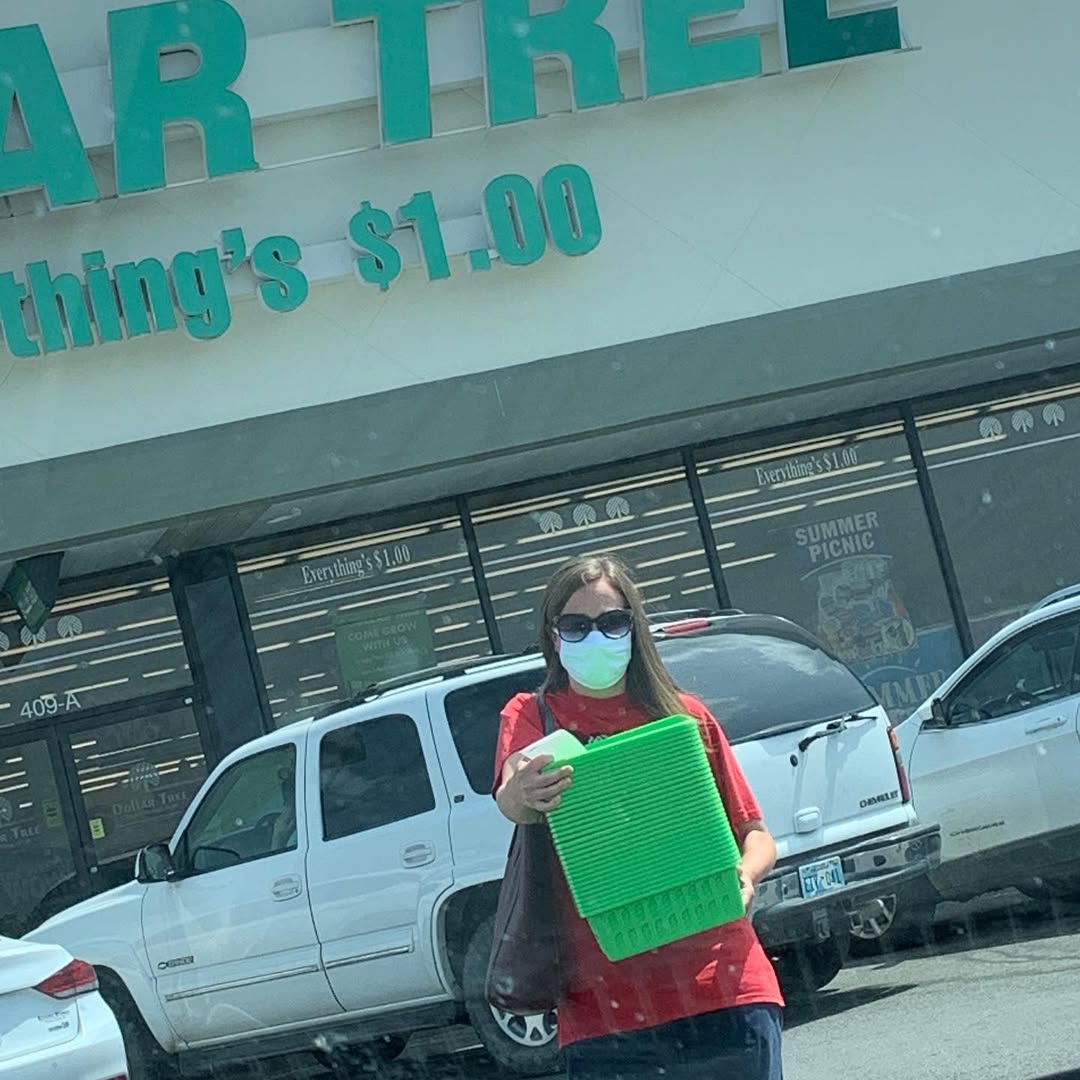}{
3:26 PM Jul 1, 2020 from Tahlequah, OK
}
&
\tweetcase{
gloom my day \#boardwalkclosed \#fencedin \#locked \#gates \#cold \#covid\_19 \#epedemic \#truphotos
}{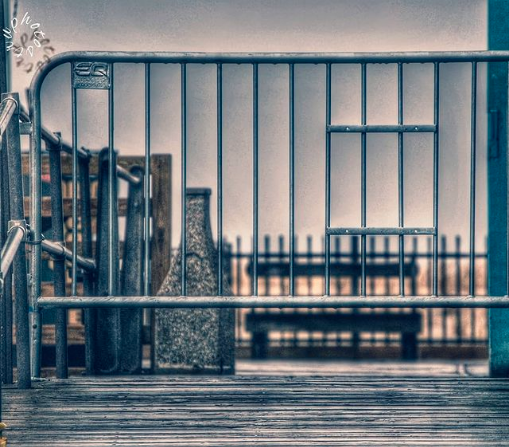}{
4:13 PM Mar 29, 2020 from Seaside Heights, NJ
}
&
\tweetcase{
Im a Barbie girl... in a Covid world \#transisbeautiful \#girlslikeus \#transvisibility \#saturdayvibes \#saturdaynight \#covid19 
}{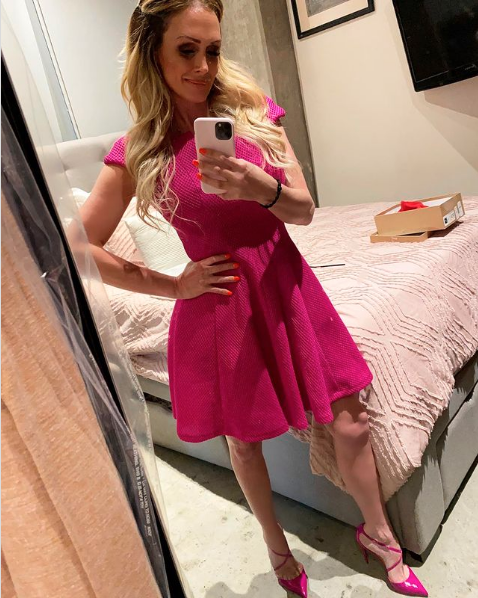}{
12:42 AM Sep 13, 2020 from Dallas, TX
}
\\
\midrule
\textbf{LLM:} \texttt{1--30 days}\par
The reason is that the tweet text explicitly uses hashtags like \#lockdown and \#coronavirus. During the COVID-19 pandemic, a quarantine period was a mandated or recommended period of isolation, typically lasting for 14 days. This falls within the 1--30 day range, making it the most logical duration for the author's stay in Manhattan.
&
\textbf{LLM:} \texttt{more than 30 days}\par
The reason is that the tweet is about buying supplies for a Vacation Bible School (\#vbs2020) at the Tahlequah United Methodist Church (@tahlequahumc). Organizing and participating in such a local church event strongly suggests that the person is a resident of Tahlequah or the surrounding area, not a temporary visitor. Therefore, they are likely to be in the city for an extended period.
&
\textbf{LLM:} \texttt{more than 30 days}\par
The image and hashtags like \#boardwalkclosed, \#locked, and \#fencedin show that public areas were shut down. During this period, travel was heavily restricted, and people were generally under stay-at-home orders. It is most likely that the author is a resident of Seaside Heights documenting the unusual state of their local area during the lockdown, rather than a tourist visiting a closed-down town.
&
\textbf{LLM:} \texttt{1--30 days}\par
The tweet includes the hashtags \#saturdayvibes and \#saturdaynight, indicating that the author is dressed up for a specific weekend event. This, combined with being in a specific city ``Dallas'', suggests that she is likely on a short trip, probably for the weekend. A weekend trip typically lasts for a few days, which falls into the 1--30 day category.
\\

\midrule

\textbf{Human:} \texttt{more than 30 days}
&
\textbf{Human:} \texttt{0--24 hours}
&
\textbf{Human:} \texttt{0--24 hours}
&
\textbf{Human:} \texttt{more than 30 days}
\\

\bottomrule
\end{tabularx}
\end{table*}

\begin{table}[h]
    \centering
    \caption{Prediction Pair Analysis. `L' indicates that the LLM's prediction is the final decision, `H' indicates that the human annotator's prediction is the final decision, and `D' indicates that the annotation was discarded.} 
    \label{tab:pair_analysis}
    \resizebox{\linewidth}{!}{%
    \begin{tabular}{@{}lccccc@{}}
        \toprule
        Prediction: <LLM, Human>  & R1 &  R2 & R3 & R4 & \# Tweets \\
        \midrule
        <0-24 hours, 1-30 days> &0 &0 &3(D) &5(D) &8 \\
        <0-24 hours, more than 30 days> &0 &20(H) &191(D) &0 & 201\\ 
        <1-30 days, 0-24 hours> &0 &0 &26(D) &24(D) & 50\\
        <1-30 days, more than 30 days> & 8(H)& 0& 0&51(D) & 59\\
        <more than 30 days, 0-24 hours> &0 & 266(L)&87(D) &0 & 353\\
        <more than 30 days, 1-30 days>& 3(L)& 0& 0&23(D) &26 \\
        \bottomrule
    \end{tabular}
    }
\end{table}

While Table~\ref{tab:pair_distribution} demonstrates baseline consensus on the diagonal, the off-diagonal cells reveal substantial discrepancies, most notably when one annotator predicts a brief stay (\texttt{0-24 hours}) while the other infers a long-term stay (\texttt{more than 30 days}). To resolve conflicts, we analyzed the labels where LLMs and human annotators disagreed and identified four main reasons for these discrepancies:
\begin{itemize}[nosep]
    \item {R1: COVID-19 quarantine duration commonsense. } LLMs or human annotators may misinterpret the typical duration in the COVID-19 quarantine period.
    \item {R2: Duration not calculated at the city level. } The annotator may overly focus on the specific location mentioned in the tweet, attempting to predict the duration based on that location, rather than at the city level.
    \item {R3: Tweet intent unclear. } The tweet might express intent to some places rather than an actual visit. 
    \item {R4: Author residency status unclear. } It’s hard to interpret whether the author is a tourist or a resident. 
\end{itemize}

The first two reasons (R1 and R2) represent the most common errors made by the LLM or human annotator, where the predictions within 30 days can be directly rejected. R3 and R4, being more ambiguous, cannot be definitively resolved, so we discarded those tweets. Examples of these cases are shown in Table~\ref{tab:system_error_merged}. To determine the final duration label, we apply the rejection and discard strategy based on the aforementioned reasoning. The remaining annotations, whether from the LLM or human annotator, are kept as the final label, while discarded annotations are excluded. The distribution of disagreement reasons for each label pair is displayed in Table~\ref{tab:pair_analysis}. In the rest of this paper, we work with these filtered tweets.




\section{Corpus Analysis}
\label{section:corpus_analysis}

\begin{figure}
    \centering
    \includegraphics[width=\columnwidth]{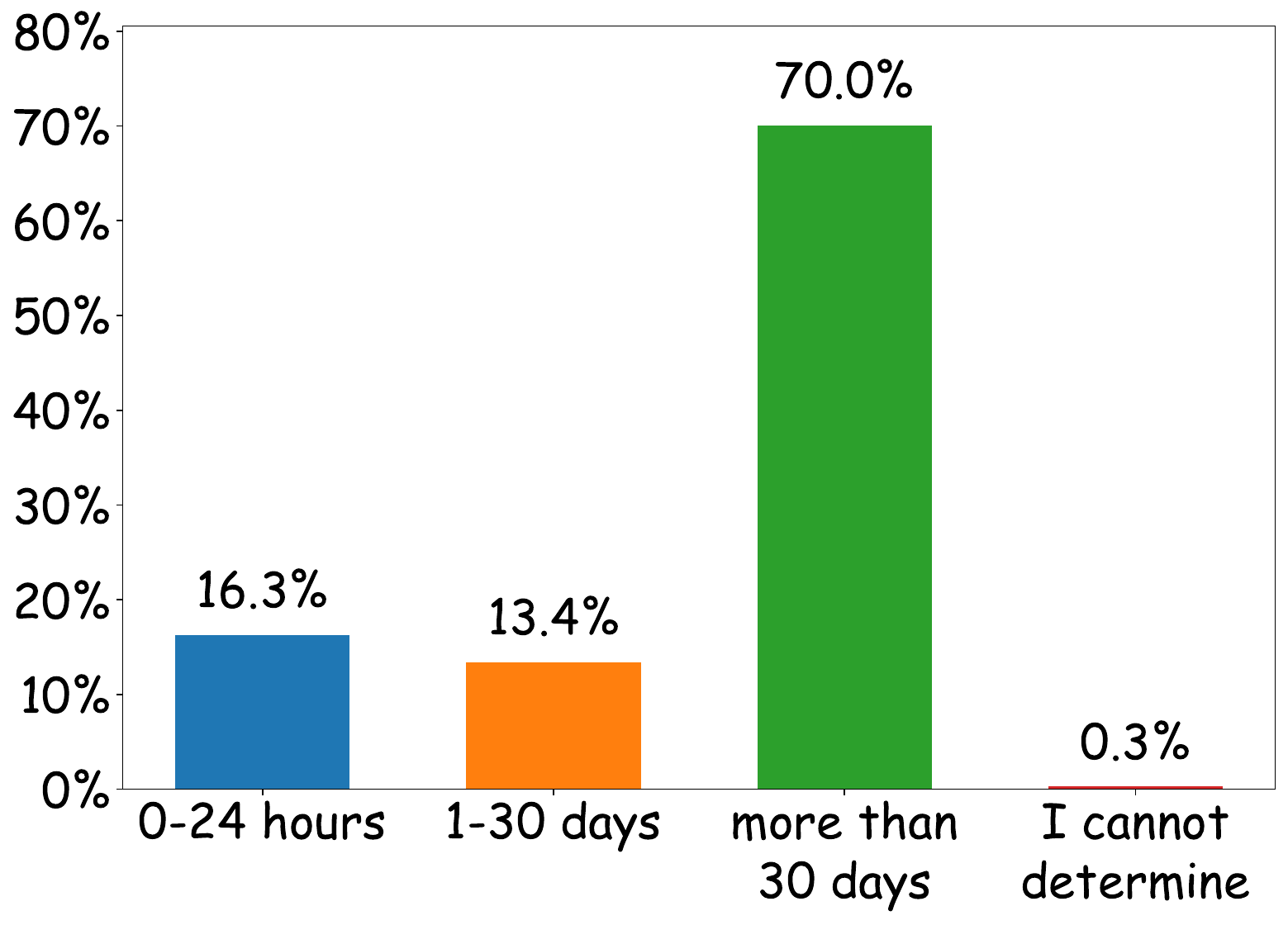}
	\caption{Label distribution. 
        }
	\label{fig:label_distribution}
\end{figure}

After human--LLM alignment and filtering (Section~\ref{section:corpus_building}), the final corpus contains 4,312 geotagged tweets with adjudicated duration labels. Figure~\ref{fig:label_distribution} shows the label distribution. Each instance pairs a tweet with its geotagged US city; we exclude the \texttt{I cannot determine} label ($<$1\% of instances) from all analyses and experiments because it does not reflect a usable duration judgment. Roughly 70\% of tweets are labeled \texttt{more than 30 days}. Labels of the remaining tweets are nearly evenly split between \texttt{1-30 days} and \texttt{0-24 hours}.

\section{Experiments and Results}
\label{section:experiments_results}
\begin{table*}[t]
\centering
\small
\setlength{\tabcolsep}{5pt}
\renewcommand{\arraystretch}{1.08}
\caption{
Weighted-average F1 scores of different model families on duration inference.
ZS-ST, FS-ST, and FS-MT denote zero-shot single-turn, few-shot single-turn, and few-shot multi-turn settings, respectively.
Text-only LLMs use tweet captions as input.
The multi-turn setting separates visual analysis and duration inference into two sequential dialogue turns. The best score within each method group is \underline{underlined}, and the best score in each evaluation setting is \textbf{boldfaced}.
A dash indicates that the setting is not applicable.
Further details are provided in Section~\ref{section:VLM} and Table~\ref{fig:prompt_examples}.
}
\label{table:results_llms}
\begin{tabular}{@{}lllcccc@{}}
\toprule
\textbf{Methods} 
& \textbf{Backbone} 
& \textbf{Input} 
& \textbf{ZS-ST} 
& \textbf{ZS-MT}
& \textbf{FS-ST} 
& \textbf{FS-MT} \\
\midrule

Human 
& Human reference 
& human judgment
& \multicolumn{3}{c}{0.37} \\

\midrule
\multirow{6}{*}{PTLM}
& RoBERTa-Large \textit{340M}        & text & 0.18 & 0.21 & -- & -- \\
& ViT-Large/16                       & image & 0.05 & 0.07 & -- & -- \\
& RoBERTa-ViT-Concatenation          & text + image & 0.22 & 0.24 & -- & -- \\
& RoBERTa-ViT-Addition               & text + image & 0.16 & 0.17 & -- & -- \\
& RoBERTa-ViT-Multiplication         & text + image & 0.13 & 0.15 & -- & -- \\
& CLIP                               & text + image & \underline{0.38} & \underline{0.38} & -- & -- \\

\midrule
\multirow{2}{*}{LLM}
& Qwen 2.5        & text& 0.28 & 0.31	&0.32	&0.35 \\
& Llama 3.2      & text& \underline{0.37} & \underline{0.40} & \underline{0.40}& \underline{0.43} \\

\midrule
\multirow{2}{*}{\shortstack[l]{VLM (w/o\\deep thinking)}}
& Claude 3 Haiku       & text + image& 0.32 & 0.35 & 0.41 & 0.46 \\
& Gemini 1.5 Flash     & text + image& \underline{0.33} & \underline{0.40} &\underline{0.40}& \underline{0.47} \\

\midrule
\multirow{6}{*}{\shortstack[l]{VLM (w/\\deep thinking)}}
& GPT5-mini        & text + image& 0.62 & 0.65	&0.68	&0.71 \\
& Grok4            & text + image& 0.70 &0.75	&0.77	&0.82 \\
& Gemini 2.5 Pro   & text + image& \textbf{\underline{0.72}} & 0.76	&0.78&	0.83 \\
& Qwen 2.5VL       & text + image& 0.71 & 0.79	&0.80&	0.84 \\
& Llama 3.2 Vision Instruct        & text + image& 0.65 & 0.69 &0.69 & 0.73 \\
& GLM-4V           & text + image& 0.70 & 0.75&\textbf{\underline{0.81}} & \textbf{\underline{0.86}} \\

\bottomrule
\end{tabular}
\end{table*}

Our problem can be modeled as a classification task.
The input is a (tweet, city) pair, and the output is a label indicating how long the authors of tweets will stay at the places they tag in their tweets after publishing tweets. We create stratified training and test splits (80\% / 20\%). We only use the first image if the tweet includes more than one image. 
Our models do not take user information (e.g., username and profile image) into account.
They make predictions based exclusively on the content of tweets.
We conduct all experiments on NVIDIA A100 GPUs. PTLMs are fine-tuned on the training set and then evaluated on the test set; LLMs and VLMs are evaluated on the test set without fine-tuning. The evaluation of different models is categorized into five types: \textbf{(1) Human:} The adjudicated labels by AMT. \textbf{(2) PTLM: } We evaluate pretrained encoder-based baselines using RoBERTa-Large~\cite{DBLP:journals/corr/abs-1907-11692} as the text encoder and ViT-Large/16~\cite{dosovitskiy2021an} as the image encoder. 
We test text-only and image-only variants, as well as multimodal variants that combine text and image representations by concatenation, addition, and element-wise multiplication. The resulting representation is passed through a two-layer classifier.
We also evaluate CLIP~\cite{radford2021learning} as a jointly pretrained image--text baseline.
\textbf{(3) LLM (Text Only):} We evaluate Qwen 2.5~\cite{bai2025qwen25vltechnicalreport} and Llama 3.2~\cite{grattafiori2024llama3herdmodels} using only the tweet text, image caption and the geotagged city. 
\textbf{(4) VLM (w/o deep thinking): }We use Claude 3 haiku and Gemini 1.5 Flash as proprietary VLMs on the held-out test set.  \textbf{(5) VLM(w/ deep thinking) :} We evaluate GPT-5-mini~\cite{GPT5mini}, Grok4, Gemini 2.5 Pro~\cite{gemini}, Qwen2.5-VL~\cite{bai2025qwen25vltechnicalreport}, Llama 3.2 Vision Instruct~\cite{grattafiori2024llama3herdmodels}, and GLM-4V\cite{vteam2025glm45vglm41vthinkingversatilemultimodal}. 

To bridge the gap between complex multimodal understanding and logical deduction, we incorporate Multi-turn Chain-of-Thought (CoT) prompting in our evaluations. By initially prompting the models to describe visual elements and their relation to textual cues, CoT creates an intermediate reasoning trace that explicitly grounds the final duration prediction. This motivation aligns with recent findings that breaking down complex reasoning tasks enhances LLM robustness. 

\subsection{Experiment setting for pretrained language models}
\label{section:tra}
We use RoBERTa-Large \textit{340M}~\cite{DBLP:journals/corr/abs-1907-11692} as the text encoder and Vision Transformer~\cite{dosovitskiy2021an} as the image encoder.~\footnote{We do not freeze RoBERTa and Vision Transformer during the training process. All their parameters are trainable.} For multimodal variants, we encode tweet text and the first attached image with the same backbones and fuse representations by concatenation, element-wise addition, or element-wise multiplication before classification.
The CLIP baseline uses \texttt{openai/clip-vit-base-patch32}~\cite{radford2021learning} for both modalities.
Each variant feeds the merged representation through two fully connected layers (hidden size 256, output size 3) with dropout~\cite{JMLR:v15:srivastava14a} (rate 0.2) before the final softmax. We use the pretrained and Vision Transformer released by HuggingFace~\cite{wolf-etal-2020-transformers} and conduct all experiments using Pytorch~\cite{NEURIPS2019_9015}.
We train all networks for up to 100 epochs using Adam optimizer~\cite{DBLP:journals/corr/KingmaB14}, categorical cross-entropy as the loss function, and a batch size of 16.
We stop the training process before 100 epochs if there is no improvement for 5 consecutive epochs.

\subsection{Experiment setting for large language models and vision language models}
\label{section:VLM}
\begin{table*}[t!]
\centering
\caption{Representative prompts used for different model settings. For text-only models, the image is converted into a textual description. For VLM models, the original image is provided directly. In the multiturn CoT setting, the model first interprets the image and its relation to the tweet, and then predicts the duration label.}
\label{fig:prompt_examples}
\scriptsize
\begin{tabularx}{\textwidth}{l >{\hsize=0.85\hsize\arraybackslash}X >{\hsize=0.80\hsize\arraybackslash}X >{\hsize=1.35\hsize\arraybackslash}X}
\toprule
 \textbf{Model} & \textbf{Content} & \textbf{Single-turn prompt} & \textbf{Multiturn prompt} \\
\midrule
\textbf{LLM (Text Only)} & 
\textbf{Tweet text:} Corona virus approved activities at the Glass House! \#COVID19 \#backtobasics \#puzzlesrock \#winelife \#deerfootwinery \#staycationfunwiththewife @ Wyomissing Park, Reading \par\vspace{1ex}
\textbf{Tweet image description:} an indoor home setting with a jigsaw puzzle spread across the floor, wine glasses, and a decorated Christmas tree &
\textit{System / user prompt:} You are a helpful assistant for inferring the duration of human activities. Please read the tweet and determine how long the author of the tweet will stay in \textbf{Reading} after posting this tweet. \par\vspace{1ex}
\textbf{Options:} \newline
1. 0--24 hours \newline
2. 1--30 days \newline
3. more than 30 days \par\vspace{1ex}
\textbf{Answer:} \textbf{2. 1--30 days.} The hashtag \#staycationfunwiththewife suggests a temporary leisure stay rather than permanent residence. The tweet describes home-based activities such as doing puzzles and drinking wine, which are compatible with a short staycation lasting multiple days but do not by themselves imply long-term residence. &
\textit{Round 1 prompt:} You are a helpful assistant for inferring the duration of human activities. Analyze how the tweet’s image description relates to its text and hashtags. \par\vspace{1ex}
\textbf{Answer:} The image description validates the tweet's narrative: a pandemic-safe, low-tech, couple-focused staycation. The immediate scene (puzzle + wine) likely represents a 2–4 hour evening activity, but the hashtag \#staycationfunwiththewife suggests this isn't a one-off event but a multi-day to week-long pattern of home-based leisure \par\vspace{1ex}
\textit{Round 2 prompt:} Please read the tweet and determine how long the author of the tweet will stay in \textbf{Reading} after posting this tweet. \par\vspace{1ex}
\textbf{Options:} \newline
1. 0--24 hours \newline
2. 1--30 days \newline
3. more than 30 days \par\vspace{1ex}
\textbf{Answer:} \textbf{2. 1--30 days.} The combination of the \#staycation hashtag, pandemic-era context, and home-based leisure activities strongly implies the author is settled in Reading for a short-to-medium-term local break. The most evidence-supported inference is a stay lasting between 1 and 30 days after posting. \\
\midrule
\textbf{VLM} &
\textbf{Tweet text:} Corona virus approved activities at the Glass House! \#COVID19 \#backtobasics \#puzzlesrock \#winelife \#deerfootwinery \#staycationfunwiththewife @ Wyomissing Park, Reading \par\vspace{1ex}
\textbf{Tweet image:} \newline \includegraphics[width=1in]{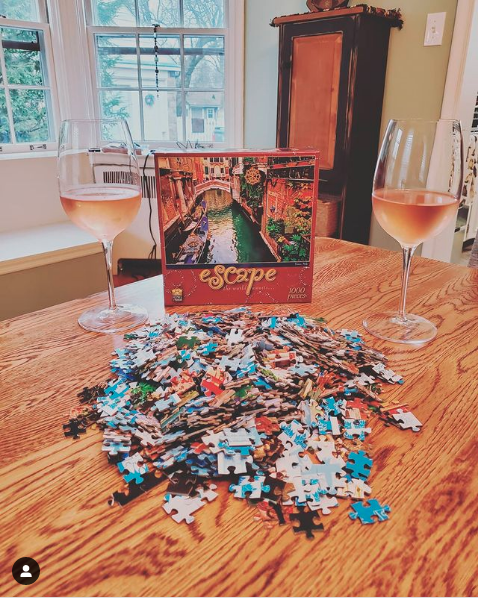} &
\textit{System / user prompt:} You are a helpful assistant for inferring the duration of human activities. Please read the tweet and determine how long the author of the tweet will stay in \textbf{Reading} after posting this tweet. \par\vspace{1ex}
\textbf{Options:} \newline
1. 0--24 hours \newline
2. 1--30 days \newline
3. more than 30 days \par\vspace{1ex}
\textbf{Answer:} \textbf{3. more than 30 days.} The tweet refers to ``Corona virus approved activities'' and includes \#COVID19, suggesting stay-at-home behavior during the pandemic. The image shows an indoor domestic setting, and \#staycationfunwiththewife further implies that the author is spending time at or near home. Taken together, these cues suggest that the author is likely a resident of Reading and will remain there for an extended period. &
\textit{Round 1 prompt:} You are a helpful assistant for inferring the duration of human activities. Describe the tweet image in detail, and analyze its relationship to the tweet text and hashtags. \par\vspace{1ex}
\textbf{Answer:} The image shows an indoor home environment with wooden flooring. A large jigsaw puzzle is spread across the floor, and wine glasses are visible nearby. These visual details align with \#puzzlesrock and \#winelife. The hashtags \#COVID19 and \#staycationfunwiththewife suggest home-based leisure during the pandemic, while \#backtobasics indicates simple indoor activities. \par\vspace{1ex}
\textit{Round 2 prompt:} Please read the tweet and determine how long the author of the tweet will stay in \textbf{Reading} after posting this tweet. \par\vspace{1ex}
\textbf{Options:} \newline
1. 0--24 hours \newline
2. 1--30 days \newline
3. more than 30 days \par\vspace{1ex}
\textbf{Answer:} \textbf{3. more than 30 days.} The tweet and image together indicate a home-based staycation during COVID-19. Because the activity appears to take place in the author's home environment, the author is likely staying in Reading long-term rather than visiting briefly. \\
\bottomrule
 \end{tabularx}
\end{table*}

To analyze the capability of recent large models (with at least 1 billion parameters) on duration commonsense understanding, we conduct extensive experiments with modern large language models (LLMs) and vision language models (VLMs).
To ensure broad coverage, we evaluate both text-only LLMs (Qwen 2.5~\cite{bai2025qwen25vltechnicalreport} and Llama 3.2~\cite{grattafiori2024llama3herdmodels}) and recent VLMs. The VLMs include open models (Qwen 2.5VL, Llama 3.2 Vision Instruct, and GLM-4V~\cite{vteam2025glm45vglm41vthinkingversatilemultimodal}) and closed models (Claude 3 Haiku, Gemini 1.5 Flash, GPT-5-mini~\cite{GPT5mini}, Grok4~\cite{grok}, and Gemini 2.5 Pro~\cite{gemini}).

To address the challenge of complex visual-linguistic alignment, we conduct experiments using single-turn and multi-turn Chain-of-Thought (CoT) prompting~\cite{ni2024visualo1understandingambiguousinstructions}. This technique has proven effective in bridging reasoning gaps in multimodal tasks. As shown in the following prompt template, the multi-turn setting separates visual analysis and duration inference into two sequential dialogue turns:

\begin{tcolorbox}[
    colback=white,
    colframe=black!75,
    boxrule=1pt,
    arc=2mm,
    auto outer arc,
    left=8pt, right=8pt, top=8pt, bottom=8pt
]
\textbf{The first round of dialogue}\\
Describe the tweet image in detail, and analyze its relationship with the tweet's text and hashtags. \\[1em]
\textbf{The second round of dialogue}\\
\{question with user prompt\}
\end{tcolorbox}
For each exemplar used in the few-shot experiments, we provide a prompt and response that systematically walk through how the visual and textual hints lead to the final duration inference. Table~\ref{fig:prompt_examples} provides a detailed example of this evaluation process.

\subsection{Results}
We benchmark the duration inference capabilities across all model families, from unimodal baselines to state-of-the-art multimodal reasoning models. Table~\ref{table:results_llms} presents the comprehensive evaluation results.

\noindent
\textbf{Effect of multi-modality.} 
For PTLM baselines, combining text and images improves performance. Specifically, CLIP outperforms models pretrained on only text or images, indicating that joint vision-language pretraining effectively bridges the modality gap. 
For LLMs, the Llama 3.2 model performs slightly better than some non-deep-thinking VLMs in the zero-shot setting. However, its reliance on generated image captions creates a bottleneck, preventing it from matching the accuracy of advanced deep-thinking VLMs. The reason could be that it struggles to fully interpret nuanced semantic duration information that requires direct visual grounding.

\noindent
\textbf{Effect of in-context learning and Multi-turn CoT.} 
Transitioning from the zero-shot to the few-shot setting consistently improves performance across virtually all models, highlighting the importance of exemplars for complex duration reasoning. For instance, the F1 score of Gemini 1.5 Flash increases from 0.33 (zero-shot single-turn) to 0.47 (few-shot multi-turn). Furthermore, Multi-turn Chain-of-Thought (CoT) prompting provides an additional performance boost. GLM-4V achieves the highest overall F1 score of 0.86 in the few-shot with Multi-turn CoT setting. Guiding the model to explicitly verbalize its visual and semantic observations before predicting the duration significantly reduces reasoning errors.

\noindent
\textbf{Bottleneck of deep-thinking models.} 
Models equipped with advanced reasoning capabilities (e.g., Qwen 2.5VL, GLM-4V, Grok4, GPT-5-mini, and Gemini 2.5 Pro) substantially outperform standard proprietary models (e.g., Claude 3 Haiku and Gemini 1.5 Flash). Despite this, F1 scores plateau around 0.86 even with few-shot exemplars and Multi-turn CoT. Current models continue to encounter fundamental limitations when processing highly ambiguous or culturally nuanced commonsense cues, as detailed in Section~\ref{section:error_analysis}.




\begin{table*}[t]
\centering
\small
\setlength{\tabcolsep}{5pt}
\renewcommand{\arraystretch}{1.08}
\caption{Error analysis of all evaluated models}
\begin{tabularx}{\textwidth}{@{}YYY@{}}
\toprule
\textbf{Context} & \textbf{Visual} & \textbf{Semantic} \\
\midrule

\tweetcase{
I carried the memories of my dad on a Social Distancing (SoDi) Half Marathon for Father's Day. \#MNStrong \#AllOutin2020 \#Minnesota \#FathersDay \#fitgay
}{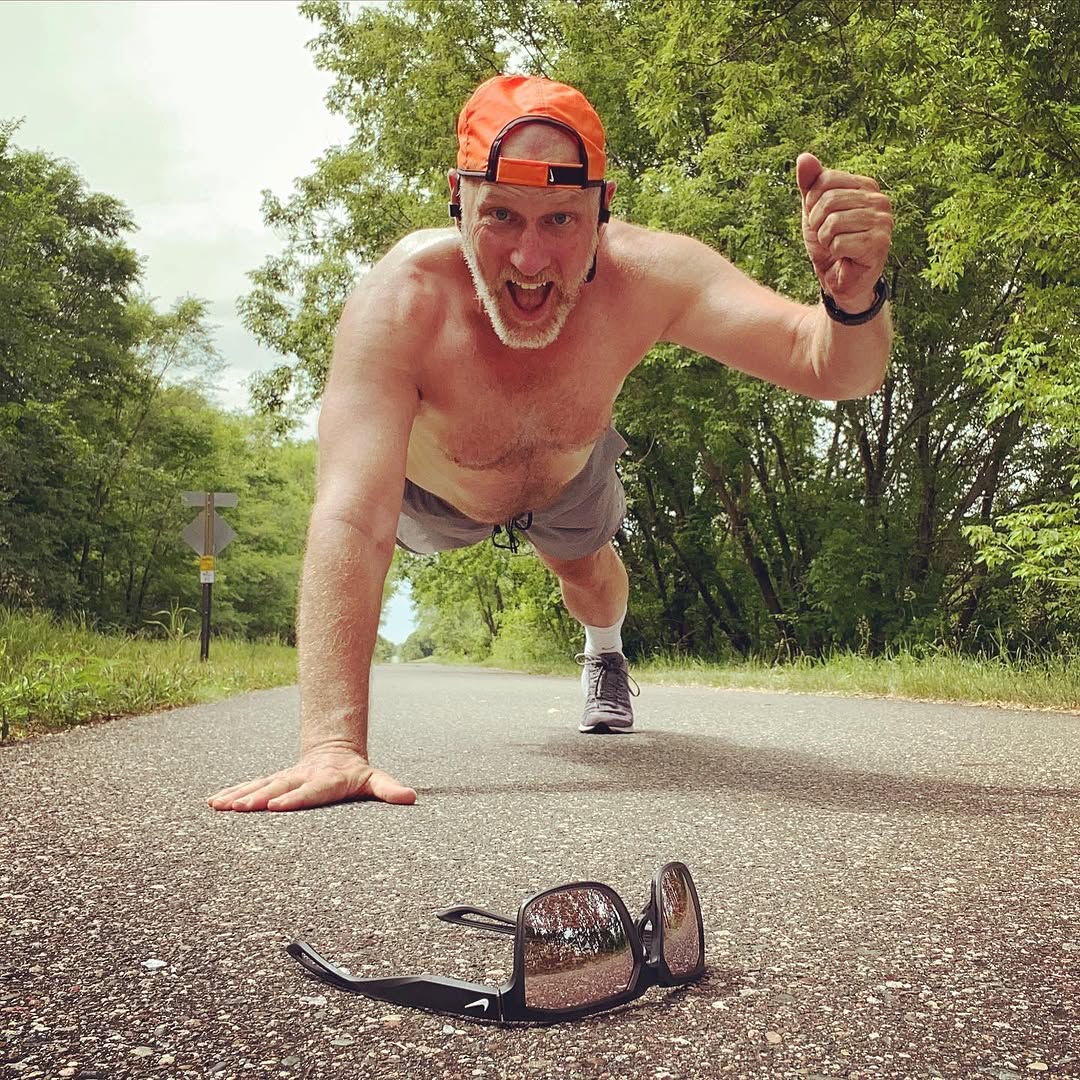}{
3:12 AM Jun 23, 2020 from St Joseph, MN
}
&
\tweetcase{
Love me a good day in \#oceancitymaryland. The weather was perfect for a walk. Most folks were \#socialdistancing. Just wish \#oceangallery was open. The \#clouds and sky were amazing.
}{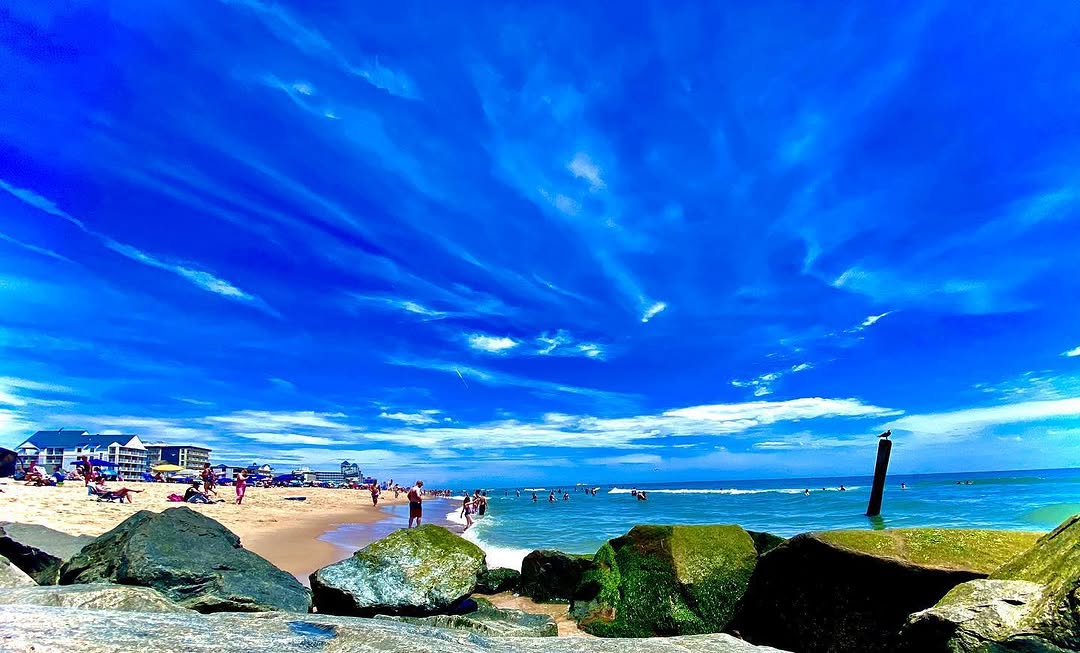}{
12:35 AM Jun 21, 2020 from Ocean City, MD
}
&
\tweetcase{
Glad to see one of my best friends Raul from Chicago and college in town! 
\#LewisUniversity \#Chicago \#Flyers \#Botines \#Liverpool 
@ Corona del Mar State Beach
}{images/semantic.jpg}{
12:29 AM Aug 18, 2020 from Newport Beach, CA
}
\\

\midrule

\textbf{Generated:} \texttt{0--24 hours}\par
The image and the tweet describe the author completing a half marathon run on a trail for Father's Day. This is a specific, time-bound activity that typically lasts a few hours. After finishing the run and posting the photo, it is most likely that the author would return home. The nature of the activity does not suggest a multi-day vacation or residency.
&
\textbf{Generated:} \texttt{0--24 hours}\par
The tweet states, "Just wish \#oceangallery was open". This implies that the author had intended to visit the Ocean Gallery during their current stay but couldn't because it was closed. The sentiment of "just wish" suggests a missed opportunity that they won't be able to fulfill on this trip, strongly indicating that their stay in Ocean City is 0 hours.
&
\textbf{Generated:} \texttt{1--30 days}\par
The reason is that the tweet mentions the author is meeting his friend from Chicago who is 
``in town''. A friend visiting from another state usually stays for a few days or a short 
vacation, which falls within the 1--30 day range. It is unlikely to be a visit for less than 
24 hours due to the travel distance, and there is no indication of a long-term stay of over 
30 days.
\\

\midrule

\textbf{Gold:} \texttt{more than 30 days}\par
The reason is that the tweet indicates the author is running a half marathon in Minnesota, with hashtags like \#MNStrong and \#Minnesota, suggesting he is a local resident. Therefore, it is highly probable that he lives in or near St. Joseph and will remain there for an extended period.
&
\textbf{Gold:} \texttt{1--30 days}\par
The reason is that the tweet mentions having "a good day in \#oceancitymaryland" and wishing a local gallery was open. The image shows that weather was good for walking, most people were keeping distance from each other. This suggest that the author is on a vacation or a short trip, which usually lasts from a few days to a couple of weeks.
&
\textbf{Gold:} \texttt{more than 30 days}\par
The tweet states, ``Glad to see one of my best friends Raul from Chicago and college in 
town!'' The phrase ``in town'' implies that Raul is visiting the city where the author is 
located. This suggests that the author either resides in Newport Beach/Corona del Mar or has 
a prolonged presence there. If the author lives in the city, their stay is continuous and 
indefinite, which falls under ``more than 30 days''.
\\

\midrule

\textbf{Analysis:} 
Fails to recognize the hashtag \#MNStrong, which indicates he is a local resident.
&
\textbf{Analysis:} 
Fails to distinguish the actual beach walk, which is visible in the image, from the unfulfilled Ocean Gallery visit.
&
\textbf{Analysis:} 
Correctly identifies the scenario as the author hosting a friend from another state, but 
incorrectly reports the friend's duration of stay as the final answer.
\\

\bottomrule
\end{tabularx}
\label{tab:detail_error}
\end{table*}
\subsection{Error Analysis}
\label{section:error_analysis}
We systematically analyzed errors across all evaluated models to determine how input influences duration interpretation. Table~\ref{tab:detail_error} provides representative examples; we group failures into three patterns.
\begin{inparaenum}[(1\upshape)]
\item \textbf{Context:} Models ignored contextual cues, such as hashtags or geotags. In the first example, failing to recognize the hashtags which indicates local residency cause the model to underestimate duration based on the activity alone.
\item \textbf{Visual:} Models overlooked subtle visual elements. In the second example, the model predicted \texttt{0--24 hours} from the wish that \#oceangallery was open, missing the beach walk in the image.
\item \textbf{Semantic:} During complex reasoning, models drifted from established facts. In the third example, even when correctly identifying a scenario (hosting an out-of-town friend), the model conflated the visitor's stay duration with the user's residency status.
\end{inparaenum} Table~\ref{tab:detail_error} provides representative examples.
\begin{figure}[htbp]
    \centering
    \includegraphics[width=\linewidth]{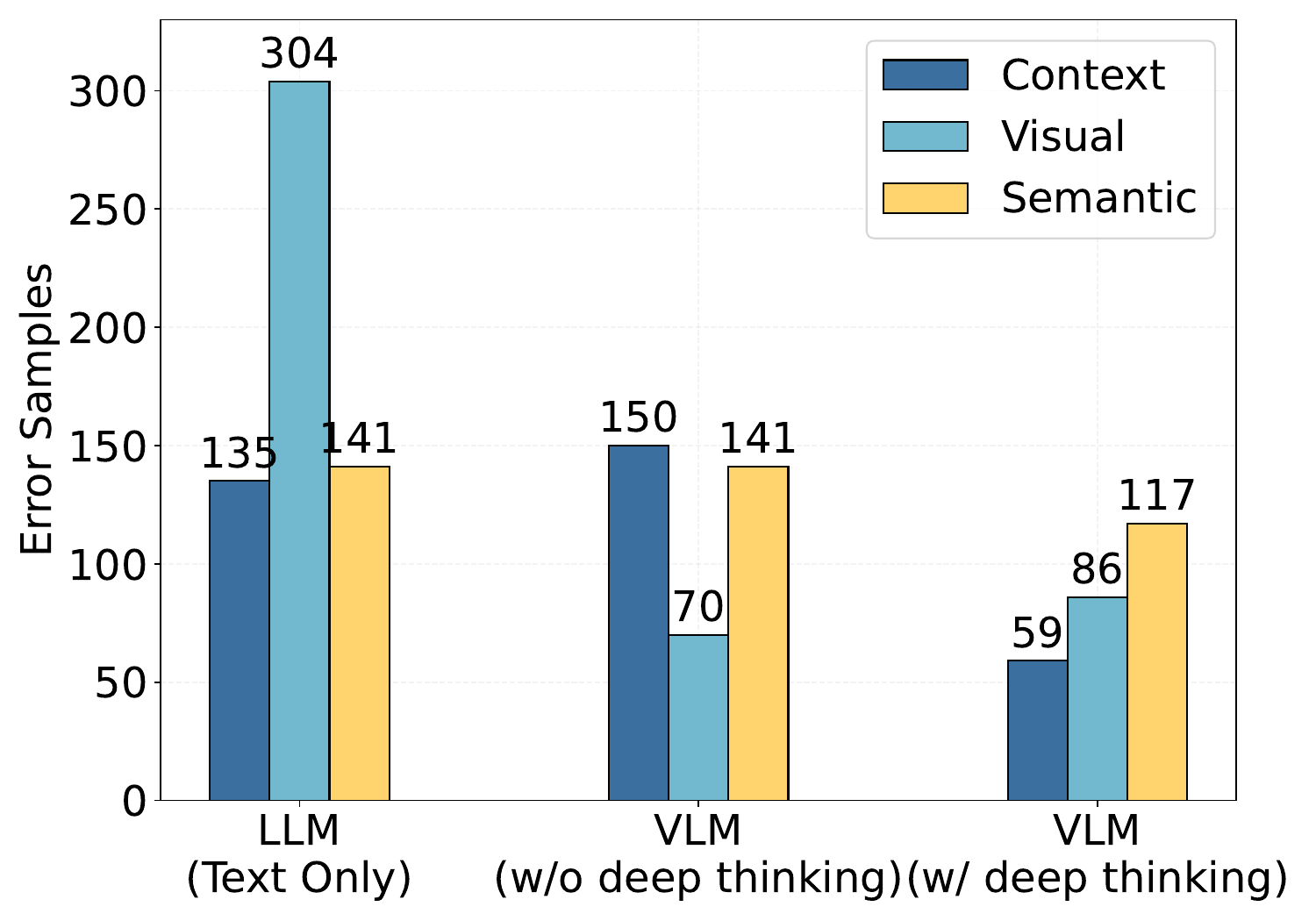}
    \caption{Distribution of error types across model categories}
    \label{fig:error}
\end{figure}

Figure~\ref{fig:error} details error distributions across model architectures. Visual errors dominated LLM (Text Only) models, reflecting their lack of visual grounding. VLM (w/o deep thinking) models resolved many visual errors but primarily struggled with context errors. The reason could be that although they perceive visual features, they lack the reasoning capacity to synthesize those visuals with disjoint textual hints like geotags or specific hashtags. Furthermore, semantic errors do not decrease with using non-deep-thinking VLMs compared to text-only LLMs. While VLM (w/ deep thinking) models minimized contextual and visual mistakes, their semantic errors remained high. Therefore, semantic reasoning remains the primary bottleneck for duration inference.

\section{Discussion}
\label{section:discussion}
While this study establishes a strong foundation for multimodal duration commonsense understanding, several important aspects regarding our methodology and its implications warrant further discussion.

\paragraph{Corpus Limitations}
Our corpus only contains tweets with geotags at the US-city level. While accounting for any location worldwide is outside the scope of this paper,
it is possible that the corpus does not generalize to other locales.
Additionally, the corpus is primarily collected from X and only contains Covid-related tweets. However, our approaches and analyses mainly focus on the relationship between location and stay duration of human activities and are not dependent on COVID-19. We believe the findings will likely generalize to other domains.

\paragraph{Annotation Challenges}
Our labels are based on inferred duration commonsense rather than directly observed future movement. Tweets often express sparse, ambiguous, or indirect intent, making fine-grained duration annotation unreliable. We therefore use three coarse labels: \texttt{0--24 hours}, \texttt{1--30 days}, and \texttt{more than 30 days}, corresponding roughly to leaving soon, temporarily staying, and long-term residence. 
The human--LLM co-annotation process improves consistency, but it does not eliminate annotation bias. Duration cues in tweets often depend on subjective interpretation, background knowledge, and assumptions about the author’s activity or residency status. In some cases, no single label is clearly correct. Future work could investigate more objective annotation protocols, collect longitudinal evidence when available, and evaluate whether finer-grained or uncertainty-aware labels better capture duration-related commonsense.

\paragraph{Ethical Considerations}
We take several steps to reduce privacy risks associated with using social media data. First, the released corpus will exclude direct user identifiers, including usernames, profile images, and tweet IDs. 
Second, we provide a takedown request process through an online form, following \cite{mirowski2019streetlearn}, so that individuals can request removal of specific examples from the corpus. We also protect crowdworker privacy. Demographic information is self-reported during the qualification task only to help us understand potential sources of annotation bias. We do not release individual-level demographic information. Access to raw demographic responses is restricted to the research team, and any demographic statistics are reported only in aggregate form to avoid identifying individual workers. Crowdworkers are compensated through Amazon Mechanical Turk according to the platform’s norms and requirements.

\section{Conclusions}
\label{section:conclusion}

We studied duration commonsense understanding. Specifically, we introduced a task: determining how long tweet authors stay at tagged locations.
To support this task, we created a corpus of 4,312 geotagged tweets. Corpus analysis shows that inferring duration requires extensive commonsense reasoning; for example, a flight departure implies a stay of hours, while visiting the Gateway Arch implies days. To the best of our knowledge, we are the first to tackle this challenging problem. Our experiments show that accurate inference requires processing highly ambiguous, culturally nuanced commonsense cues and maintaining logical consistency across disjoint visual and textual signals. Our error analysis demonstrates that, even for deepthinking VLMs, complex semantic reasoning remains a fundamental limitation.


\bibliography{duration}
\bibliographystyle{IEEEtran}
\end{document}